\documentclass[sigconf]{acmart}

\renewcommand\footnotetextcopyrightpermission[1]{} 
\def\BibTeX{{\rm B\kern-.05em{\sc i\kern-.025em b}\kern-.08emT\kern-.1667em\lower.7ex\hbox{E}\kern-.125emX}}
    
\setcopyright{none}

\usepackage{array}
\newcolumntype{L}[1]{>{\raggedright\let\newline\\\arraybackslash\hspace{0pt}}m{#1}}
\newcolumntype{C}[1]{>{\centering\let\newline\\\arraybackslash\hspace{0pt}}m{#1}}
\newcolumntype{R}[1]{>{\raggedleft\let\newline\\\arraybackslash\hspace{0pt}}m{#1}}

\begin{document}

%
\title{FCC-GAN: A Fully Connected and Convolutional Net Architecture for GANs}
%
\author{Sukarna Barua}
\affiliation{%
  \institution{The University of Melbourne}
  \city{Victoria-3010}
  \state{Australia}
}

\author{Sarah Monazam Erfani}
\affiliation{%
  \institution{The University of Melbourne}
  \city{Victoria-3010}
  \state{Australia}
}
\author{James Bailey}
\affiliation{%
  \institution{The University of Melbourne}
  \city{Victoria-3010}
  \state{Australia}
}


 




%

%
\begin{abstract}
Generative Adversarial Networks (GANs) are a powerful class of generative models. Despite their successes, the most appropriate choice of a GAN network architecture is still not well understood. GAN models for image synthesis have adopted a deep convolutional network architecture, which eliminates or minimizes the use of fully connected and pooling layers in favor of convolution layers in the generator and discriminator of GANs. In this paper, we demonstrate that a convolution network architecture utilizing deep fully connected layers and pooling layers can be more effective than the traditional convolution-only architecture, and we propose FCC-GAN, a fully connected and convolutional GAN architecture. Models based on our FCC-GAN architecture learn both faster than the conventional architecture and also generate higher quality of samples. We demonstrate the effectiveness and stability of our approach across four popular image datasets.
\end{abstract}
%
%
%
\keywords{Generative adversarial networks, GAN architecture, fully connected layers}

%

%
\maketitle

\section{Introduction}
Generative Adversarial Networks (GANs) have attracted wide attention as powerful generative models. A GAN model consists of a generator network that generates fake samples from random noise vectors, and a discriminator network that provides probabilistic feedback about how well the generated samples resemble a real distribution. Initially formulated by~\cite{gan}, GANs have been successfully used for image generation~\cite{laplaciangan,dcgan}, image in-painting~\cite{imageinpainting}, image super-resolution~\cite{imgsuperres} and text-to-image conversion~\cite{text2image}. 

Despite their attractive theoretical framework, it can be difficult to train GANs in practice, due to instability issues such as mode collapse and vanishing gradients~\cite{gan,wgan}. There has been considerable work investigating the stabilization of GAN training via the use of different network architectures~\cite{dcgan,laplaciangan,ebgan,unrolledgan} and objective functions~\cite{wgan,improvedwgan,doubleimprovedwgan}. However, the interaction between the underlying network architecture and the sample generation process of GANs is still not well understood. {\em Indeed, the choice of what architectures are most appropriate to use for GANs remains an open problem.}

Existing research usually deploys Deep Convolutional Networks (CNNs) for image synthesis tasks in GANs. Specifically, in traditional architectures, convolution layers are mostly used, and fully connected and pooling layers are eliminated or minimized~\cite{dcgan,wgan,improvedwgan,sngan}. Such practices have been largely inspired by the DCGAN model~\cite{dcgan}, which achieved significant image quality and training stability through its adoption of the all convolutional~\citep{allconvnet} architecture. A set of transposed convolution layers are used in the generator for upsampling, while strided-convolution layers are used in the discriminator to downsample images. For high resolution image synthesis, convolution networks based on ResNet~\citep{resnet} blocks instead of simple convolution layers are employed~\citep{improvedwgan,acgan,sngan}. The final network architectures are usually formed with only a single fully connected layer in the generator (to receive input noise) and discriminator (to produce the output probability). Whilst the use of such convolution-only architecture has been effective for delivering high training stability and sample quality for GANs, in this paper {\em we demonstrate that the use of multiple fully connected layers together with convolution layers can perform even better than the conventional architecture}. Thus, we propose FCC-GAN, a fully connected and convolutional net architecture for GANs.

Conventional GAN generators approach image generation as a single process achieved by a deep convolutional network. In contrast, our philosophy is to model image generation as two separate tasks: (i) conversion of the low-dimensional input noise to a high-dimensional intermediate representation of image features, and (ii) conversion of these features to an output image. An FCC-GAN generator uses deep fully connected layers for the first task and convolution layers for the second. The fully connected network used for mapping noise vectors to image features learns a consistent relationship between different noise vectors and image features leading to more natural images. Indeed our premise is that convolution layers are not well suited for such global mapping operations, due to an over emphasis on local connectivity enforced by shared-weights.

Similarly, FCC-GAN employs deep fully connected layers after a series of convolution layers in the discriminator. In conventional discriminators, convolution layers extract high-dimensional image features which are directly passed to an output layer for classification. In contrast, an FCC-GAN discriminator uses the deep fully connected network to map the high-dimensional features to a lower-dimensional space before final classification. We demonstrate that discrimination in this lower-dimensional space can prevent the discriminator loss from becoming too low. A higher loss for the discriminator results in larger gradients and faster learning for the generator. Finally, we show that the use of pooling in the discriminator is more effective than the conventional choice of strided-convolution layers for our FCC-GAN architecture. 

To summarize, we investigate and revisit two widely adopted architectural assumptions made in  GANs: (i) use of  convolution-only architecture minimizing the number of fully connected layers to just one in both the generator and discriminator, and (ii) use of strided-convolution replacing pooling layers. We demonstrate that these conventions are not necessarily the best choices and that more effective architectures are possible by combining {\em deep fully connected and pooling layers} together with convolution layers. Our proposed FCC-GAN architecture demonstrates significant improvements over conventional convolution models in terms of sample quality, learning speed, and training stability. Our contribution can be summarized as follows:

\begin{itemize}
\item We propose FCC-GAN, an architecture consisting of deep fully connected and convolution layers for both the generator and discriminator in GANs. Our proposed architecture generates higher quality samples than conventional architectures on a variety of benchmark image datasets.

\item We demonstrate that FCC-GAN has a faster learning curve than the conventional architecture, and can produce recognizable good quality images after just a few epochs of training.

\item Models based on FCC-GAN architecture perform better than those of the conventional CNN architecture on the benchmark datasets in term of Inception score and Fr\'etchet Inception Distance.

\item Our proposed architecture exhibits higher stability than the conventional architecture, and is able to avoid mode collapse for a wide variety of experimental settings.
 
\end{itemize}

\section{Related Work}

In this section, we briefly review the existing literature on GANs. At the end of this section, we describe how our work is different from existing research.

GANs were first formulated in~\cite{gan}, which demonstrated their potential as a generative model. GANs became popular for image synthesis based on successful use of deep convolution layers~\cite{dcgan,laplaciangan}. The generator in the standard model maps a single noise vector to an output distribution~\cite{gan,dcgan}. Additional information in the form of latent codes can be combined with the noise vector to control output attributes and improve the sample quality~\cite{conditionalgan,acgan,infogan,catgan}. For example, class labels combined with noise vectors as input to the generator can generate supervised data~\cite{conditionalgan}. Side-information such as image captions and bounding box localization can also be combined with class information to improve the quality of images~\cite{conimgsyn,reedwwdraw}. Maximizing mutual information between the input latent variables and the GAN outputs can produce a disentangled and interpretable representation~\cite{infogan}. In semi-supervised GAN models, the discriminator is trained to predict the correct label for each real sample, in addition to discriminating real and fake data~\cite{acgan,ganimproved,odenasemigan}. Such models produce better image quality than their unsupervised counterparts~\cite{ganimproved,acgan}.  

Optimizing the standard GAN objective function was shown to be similar to minimizing the Jensen Shannon (JS) divergence~\cite{gan} between the real distribution and generator's distribution. Moreover, the traditional optimization process of GANs is a special case of more general variational divergence estimation~\cite{varestimation}. Minimizing the JS divergence of two distributions, which have non-overlapping support can result in vanishing gradients hurting the training of GANs~\cite{wgan}. A loss function based on Wasserstein distance (WGAN) was introduced to improve stability~\cite{wgan}. WGAN training was further improved by penalizing the gradients in the loss function~\cite{improvedwgan} and adding a consistency term~\cite{doubleimprovedwgan}. GANs can also be trained using auto-encoder based discriminators~\cite{ebgan}, various types of f-divergences~\cite{fgan}, and other objective functions such as Loss-Sensitive GAN~\cite{lsgan} and Least Squares GAN~\cite{leastsquaregan}. Weight normalization in the discriminator~\citep{sngan} and large batch sizes~\citep{largescalegan} has been shown to be very useful in high resolution image synthesis.

However, most existing works on image synthesis have adopted the conventional deep convolution architecture in the generator and discriminator of GANs (not using or minimizing the use of fully connection layers). In this aspect, our work is different since we investigate an architecture composed of both deep fully connected and convolution layers. {\em To the best of our knowledge, our work is the first to demonstrate a network combining convolution layers with three or more fully connected layers is effective for GANs.}

\begin{figure}[ht!]
\centering
\begin{tabular}{cc}
    \multicolumn{2}{c}{\includegraphics[width=65mm]{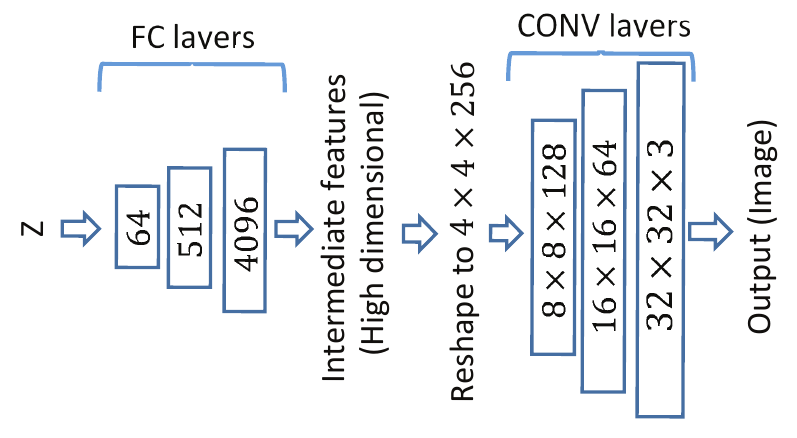}} \\
    \multicolumn{2}{c}{(a) FCC-GAN generator} \\
    \multicolumn{2}{c}{\includegraphics[width=75mm]{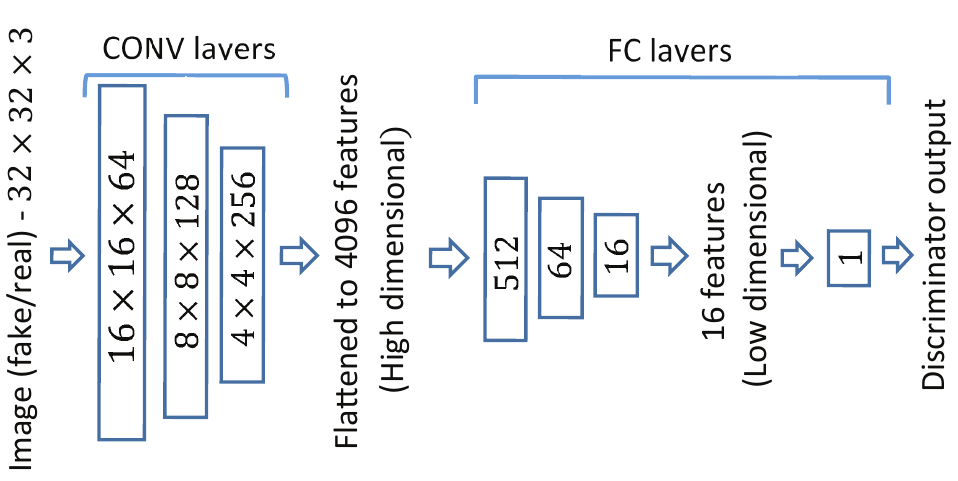}} \\
    \multicolumn{2}{c}{(b) FCC-GAN discriminator} \\
	\includegraphics[width=32mm]{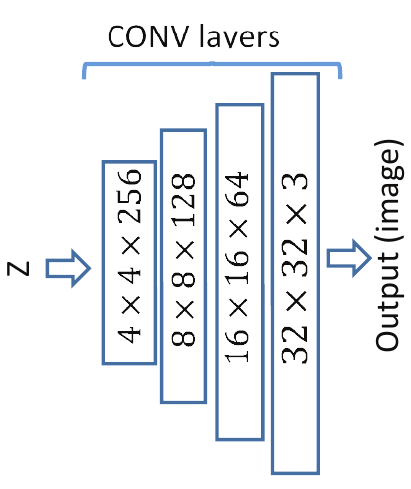} & \includegraphics[width=34mm]{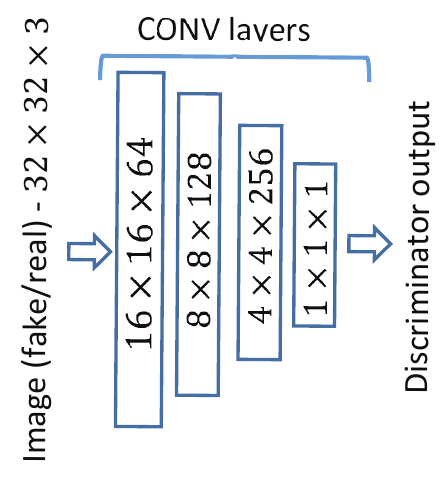} \\
	{\small (c) Conventional generator} & {\small (d) Conventional discriminator} \\
\end{tabular}
\caption{A simple example of generator and discriminator networks in the proposed FCC-GAN architecture (a-b) and conventional CNN architecture (c-d). The models generate $32\times32\times3$ RGB images from random noise vector $z$. Numbers inside the boxes denote the output shape of the layer (for CONV layers) or number of nodes (for FC layers). Note that the first (last) layer of the generator (discriminator) in conventional architecture can be considered a fully connected layer, which is employed at the input (output). For high resolution image synthesis, simple convolution layers are usually replaced with ResNet~\citep{resnet} blocks.}
\label{fig:hybridmodel}
\end{figure}

\section{FCC-GAN: A Fully Connected and Convolutonal GAN Architecture}
We propose two simple modifications to the conventional principles of GAN architecture by (i) incorporating multiple fully connected layers in both the GAN generator and discriminator networks and (ii) using pooling layers with unit stride convolution in place of strided convolution layers in the discriminator. Figures~\ref{fig:hybridmodel}(a-b) visualizes an example model of our proposed FCC-GAN architecture for generating $32 \times 32 \times 3$ images. The generator consists of two parts: a series of fully connected (FC) layers, followed by a series of convolution (CONV) layers. The FC part receives a low-dimensional (e.g., 100) noise vector $z$ and progressively converts it to a high-dimensional (4096 features) intermediate representation of image features. This representation is then reshaped  ($4 \times 4 \times 256$) for a convolution block. A series of transposed convolution layers convert the intermediate features to an output image ($32 \times 32 \times 3$). 
The discriminator is constructed with a stack of convolution layers followed by a stack of FC layers. The convolution layers extract high-dimensional ($4 \times 4 \times 256$) features from input images ($32 \times 32 \times 3$). The extracted features are flattened and fed to an FC part which progressively maps them to a lower-dimensional space for classification by an output layer. Average pooling layers are used in the discriminator for downsampling operation which implies unit-stride convolution layers are used instead of strided convolutions. 
A comparison of FCC-GAN architecture versus the conventional CNN architecture is provided in Figures~\ref{fig:hybridmodel}(c-d). 

Specifically, we propose the following architectural modifications to the conventional deep convolution architecture:
\begin{itemize}
    \item Use of multiple deep fully connected layers before convolution layers in the generator to convert the low-dimensional noise vector to a high-dimensional representation of image features.
    \item Use of multiple deep fully connected layers in the discriminator to map the high-dimensional features extracted by convolution layers to a lower-dimensional space before classification. 
    \item Use of average pooling with unit-stride convolution in the discriminator, instead of the conventional choice of strided-convolution. 
\end{itemize}

The architecture shown in Figures~\ref{fig:hybridmodel}(a-b) can be adapted for images of different resolutions by appropriately modifying the shape and depth of the convolution stack. In this study, we demonstrate the approach on four popular image datasets of three different resolutions ($28 \times 28$, $32 \times 32$, and $64 \times 64$ pixels).

\subsection{Rationale and Benefits of the FCC-GAN Architecture}

\textbf{(i) Mapping noise to intermediate image features---}
The FC network in the generator of FCC-GAN serves three purposes. First, it learns the required non-spatial mapping from random noise vectors to intermediate image features. Second, it captures the inherent relationship between different noise vectors that should be mapped to similar features of the same class of images. Third, it address the problem of shared weights in the conventional architecture, which prevent the convolution layers from generating subtle variations in different spatial zones of the same convolution filter. These variations are needed to produce realistic images. Inclusion of initial fully connected layers facilitates  capture of these subtle variations, and enforces them in the intermediate features. This helps the generator to produce more natural images. Fully connected layers are better than convolution layers to realize these objectives because their mapping is more non-spatial and the shared weights of convolution layers restrict their capacity to learn such general transformations.

\noindent\textbf{(ii) Discrimination in a low-dimensional feature space---}
Conventional discriminators extract high-dimensional features from an input image using a series of strided-convolution layers. These features are then passed to an output layer, either a convolution layer or a fully connected node. Building a decision boundary in this high-dimensional feature space to separate real and fake data samples has two disadvantages: (1) Constructing the boundary is easier (too easy) in this high-dimensional space, therefore the discriminator loss decreases quickly to very small values. For GANs, a very small discriminator loss is harmful for training the generator since the gradients become very small or vanish completely. (2) In a high-dimensional space, the distance between the decision boundary and class regions is more likely to be large. Thus, the gradients may point to random directions and may not be very accurate to train the generator, deteriorating convergence rate. Such situations are more likely to occur at the beginning of training when the generator's distribution is completely different from the real one. 

The fully connected layers in the FCC-GAN discriminator are used to map the high-dimensional image features to a lower dimensional space before classification. Such forcing of the final output node to discriminate in a low-dimensional space serves two purposes: (1) it brings the decision boundary closer to class regions, thus making the gradient directions more accurate, and (2) it is comparatively harder to build the decision boundary in a low-dimensional space. This prevents the discriminator loss from becoming small too quickly, especially at the initial stages of training.

Our motivations for using FC layers is partly empirical (see later experiments), and partly due to the following reasons. First, having no shared-weights, FC layers are more independent to learn the required non-spatial dimensional mapping task. Second, FC layers are typically harder to train for images than convolution layers are.  Hence, they assist in preventing the low discriminator loss. Indeed a larger loss for the discriminator after the addition of FC layers is clearly evident in all of our experiments. In addition, this is complemented by a low generator loss at the beginning, suggesting that the higher loss actually helps the generator learn quickly.

\noindent\textbf{(iii) Average pooling in the discriminator---} Conventional GAN models have mostly favored the use of strided-convolution in the discriminator for downsampling images, e.g., \cite{dcgan,wgan}. As we show later in experiments, use of average pooling results in a performance boost for our FCC-GAN architecture. Average pooling adds regularization in the feature extraction process by averaging features from adjacent spatial zones. We believe that such regularization becomes particularly effective when followed by deep fully connected layers.

\begin{table}[h!]
  \begin{center}
    \caption{Generator and discriminator network used for CIFAR-10 and SVHN datasets for  CNN and FCC-GAN models. CONV($x,y,z$) is a convolution layer with filters=$x$, kernel=$y \times y$, and stride=$z$. CONVT($x,y,z$) is a transposed convolution layer with filters=$x$, kernel=$y \times y$, and stride=$z$. FC($x$) is a fully connected layer with $x$ output nodes. BN implies a batch-normalization layer and R implies reshape. For simplicity, the activation functions are now shown. The architecture of FCC-GAN model is shown with strided convolution. }
    \label{tab:hybridmodelcifar10}
    \begin{tabular}{c|c}
			\hline
      \textbf{CNN Model} & \textbf{FCC-GAN Model} \\
			\hline
			\multicolumn{2}{c}{Generator network} \\
      \hline
			Input: Z(100)&Input: Z(100)\\
			R(1,1,100)&FC(64), FC(512), FC(4096), BN\\
			CONVT(256,4,1), BN&R(4,4,256)\\
			CONVT(128,4,2), BN&CONVT(128,4,2), BN\\
			CONVT(64,4,2), BN&CONVT(64,4,2), BN\\
			CONVT(3,4,3)&CONVT(3,4,3)\\
			Output: (32, 32, 3)&Output: (32, 32,3)\\

			\hline
			\multicolumn{2}{c}{Discriminator network} \\
			\hline
			Input: (32,32,3)&Input: (32,32,3)\\
			CONV(64,4,2), BN&CONV(64,4,2), BN \\
			CONV(128,4,2), BN&CONV(128,4,2), BN \\
			CONV(256,4,2), BN&CONV(256,4,2), BN \\
			CONV(1,4,1) &FC(512), FC(64), FC(16), FC(1)\\
			Output: 1&Output: 1\\
    \end{tabular}
  \end{center}
\end{table}

\section{Experiments}
We evaluate the effectiveness of FCC-GAN architecture, comparing it with the conventional CNN architecture. We train GAN models using both architectures on four benchmark image datasets: MNIST, CIFAR-10, SVHN~\citep{svhn}, and CelebA~\citep{celeba}. To demonstrate the generalizability of FCC-GAN architecture, we test models using two different objective functions used widely in the literature: the standard GAN objective function~\cite{gan} and the Wasserstein distance from WGAN~\cite{wgan}. For this study, we only focus on unsupervised GAN training. Codes for FCC-GAN implementation can be found in \emph{\url{https://github.com/sukarnabarua/fccgan}}.

\textbf{ Experimental Settings:} We train the FCC-GAN models twice: once using strided convolution in the discriminator and once using average pooling with unit-stride convolution. This was done to demonstrate the impact of pooling. While reporting the results for both of these models, we differentiate them as FCC-GAN-S (with strided convolution) and FCC-GAN-P (with pooling). In other places, we simply use FCC-GAN to refer to the latter model with pooling. Table~\ref{tab:hybridmodelcifar10} shows the network architecture of the CNN and FCC-GAN-S models used for CIFAR-10 and SVHN datasets. For FCC-GAN-P model, we just replace each strided CONV($x,y,2$) layer of the discriminator with a CONV($x,y,1$) layer followed by an average pooling layer with pool size of $2 \times 2$.

For building the CNN architecture, we adopted the suggestions of the DCGAN and WGAN methods.  Batch-normalization~\cite{batchnorm} was used in both the generator and discriminator (see Table~\ref{tab:hybridmodelcifar10}). In all layers except the output, ReLU~\cite{relu} activation was used in the generator and LeakyReLU~\cite{leakyrelu} with a leaky slope of 0.2 in the discriminator. The generator used Tanh output whilst in the discriminator, Sigmoid was used for standard GAN training and no activation was used for WGAN training.  For MNIST dataset, we changed the kernel size of the convolution layers to match the dimension of the MNIST dataset ($28\times28$). Note that the first (last) convolution layer in the generator (discriminator) of CNN models can be considered equivalent to a fully connected layer. For further details on architecture, please see Appendix~\ref{sec:modelarchdetails}.
\begin{figure}[ht!]
	\centering
	\small
	\begin{tabular}{ccc}
		\includegraphics[width=25mm]{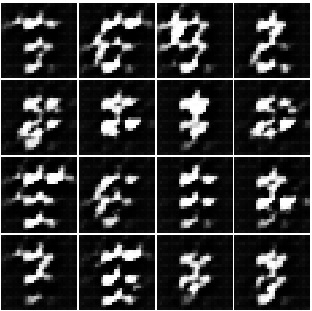} & 
			\includegraphics[width=25mm]{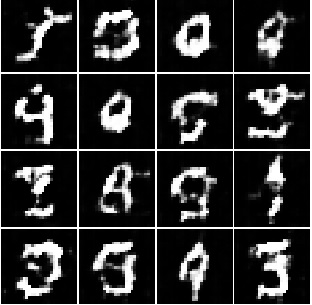} & 
				\includegraphics[width=25mm]{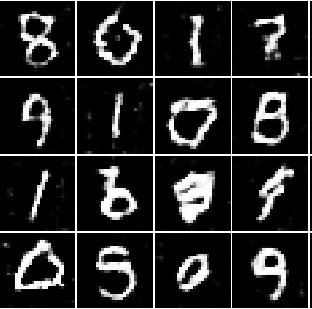} \\ 
		(a) CNN & (b) FCC-GAN-S & (c) FCC-GAN-P \\
		\includegraphics[width=25mm]{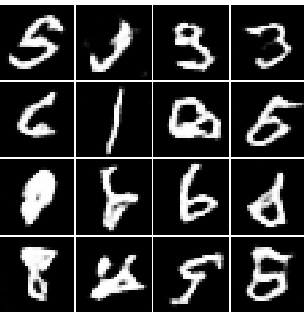} & 
			\includegraphics[width=25mm]{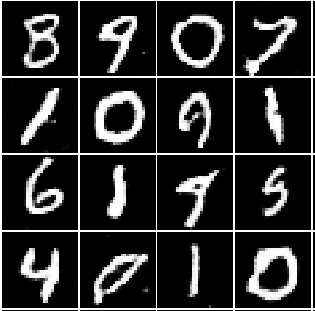} & 
				\includegraphics[width=25mm]{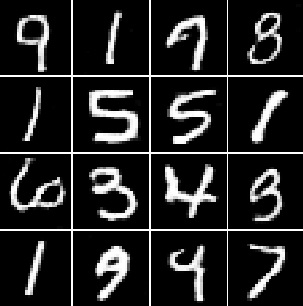} \\ 
		(d) CNN & (e) FCC-GAN-S & (f) FCC-GAN-P \\
		\includegraphics[width=25mm]{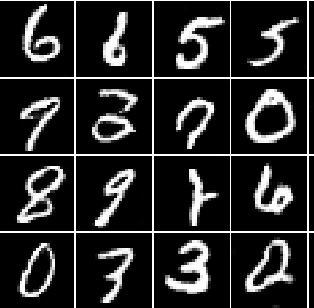} & 
			\includegraphics[width=25mm]{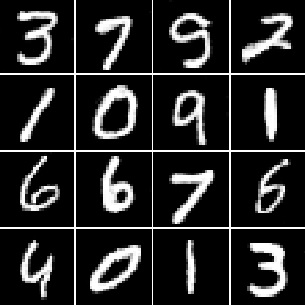} & 
				\includegraphics[width=25mm]{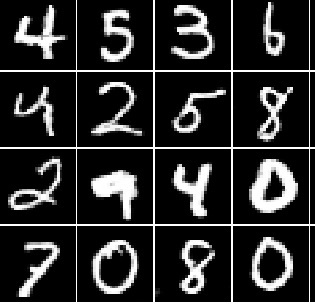} \\ 
		(g) CNN & (h) FCC-GAN-S & (i) FCC-GAN-P \\
\end{tabular}
\caption{$28\times28$ pixel MNIST images for standard GAN training. Images generated by conventional CNN and FCC-GAN models after 1 epoch (a-c), 5 epochs (d-f), and 50 epochs (g-i)}
\label{fig:dcganmnistoutput}
\end{figure}

\begin{figure}[ht!]
	\centering
	\small
	\begin{tabular}{ccc}
		\includegraphics[width=25mm]{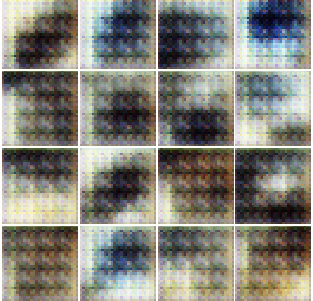} & 
			\includegraphics[width=25mm]{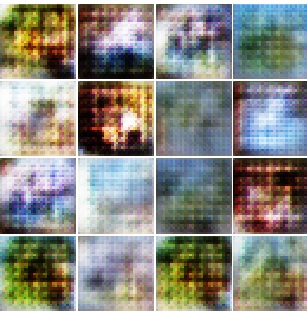} & 
				\includegraphics[width=25mm]{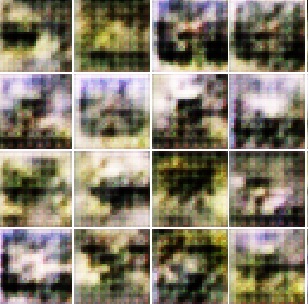} \\ 
		(a) CNN & (b) FCC-GAN-S & (c) FCC-GAN-P \\
		\includegraphics[width=25mm]{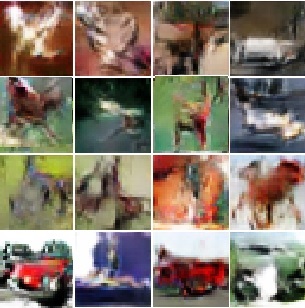} & 
			\includegraphics[width=25mm]{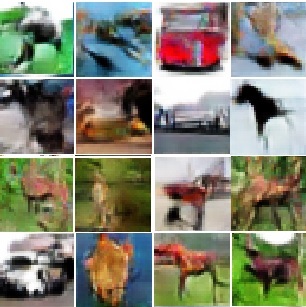} & 
				\includegraphics[width=25mm]{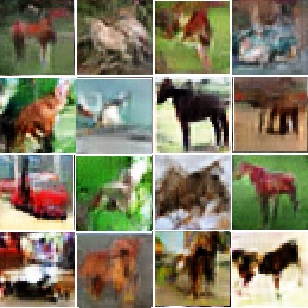} \\ 
		(d) CNN & (e) FCC-GAN-S & (f) FCC-GAN-P \\
		\includegraphics[width=25mm]{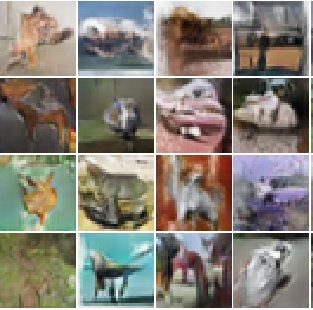} & 
			\includegraphics[width=25mm]{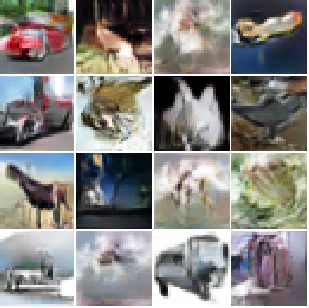} & 
				\includegraphics[width=25mm]{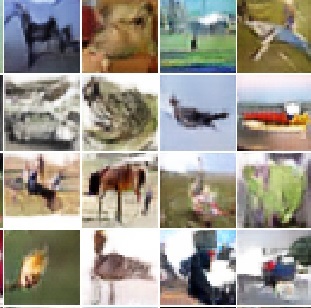} \\ 
		(g) CNN & (h) FCC-GAN-S & (i) FCC-GAN-P \\

\end{tabular}
\caption{$32\times32$ pixel CIFAR-10 images for standard GAN training. Images generated by conventional CNN and proposed FCC-GAN models after 1 epoch (a-c), 35 epochs (d-f), and 150 epochs (g-i)}
\label{fig:dcgancifar10output}
\end{figure}
For all experiments, we used the ADAM~\cite{adam} optimizer for standard GAN training and RMSProp for WGAN training, following the suggestions from DCGAN and WGAN methods. For a fair comparison, we kept all training parameters such as batch size, learning rate, and number of iterations, the same for all compared models during training. For standard GAN training, we ran a different number of iterations for different datasets: 60 epochs (approx. $94K$ iterations) on MNIST, 100 epochs (approx. $228K$ iterations) on SVHN, and 150 epochs (approx. $234K$ iterations) on CIFAR-10. The number of epochs were kept the same for WGAN training.

We compare the architectures based on the visual quality of generated samples, as well as the Inception score~\cite{ganimproved} and Fr\'etchet Inception Distance (FID)~\citep{fid}, two popular metrics for measuring visual quality and diversity of GAN outputs. We compute the Inception score for CIFAR-10 and SVHN datasets following the method proposed by~\cite{ganimproved}. For computing the score on the MNIST dataset, we followed the approach of~\cite{mnistinception}. For computing FID, we followed the approach proposed by the original authors~\citep{fid}.

\textbf{Results on Standard GAN Training:}
We summarize results of standard GAN training for the two architectures. Figure~\ref{fig:dcganmnistoutput} shows the images generated after epoch 1, 5, and 50 for the MNIST dataset. We  observe that after just one epoch, the images generated by the FCC-GAN models contain some recognizable digit structures, whereas such structures are not present for the CNN model. Both the FCC-GAN models learn the distribution much more quickly than the CNN model. After five epochs, FCC-GAN models generate clearly recognizable digits, while the CNN model does not. After epoch 50, all models generate good images, though FCC-GAN models still outperform the CNN model in terms of image quality.

\begin{figure}[ht!]
	\centering
	\small
	\begin{tabular}{ccc}
		\includegraphics[width=25mm]{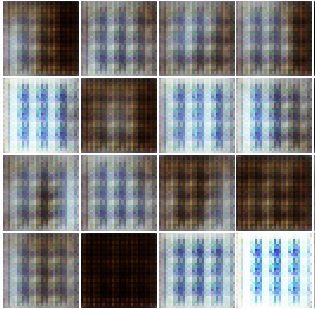} & 
			\includegraphics[width=25mm]{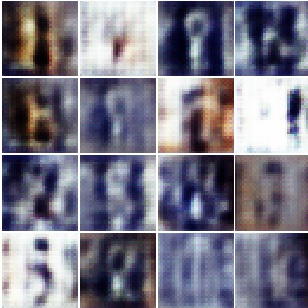} & 
				\includegraphics[width=25mm]{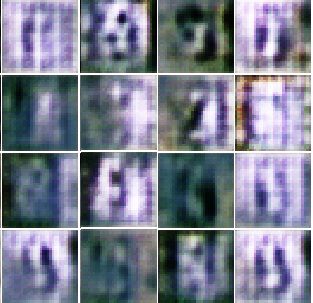} \\ 
		(a) CNN & (b) FCC-GAN-S & (c) FCC-GAN-P \\
		\includegraphics[width=25mm]{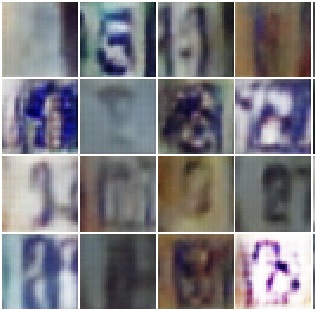} & 
			\includegraphics[width=25mm]{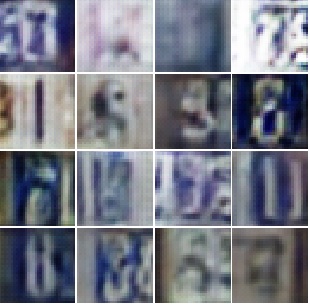} & 
				\includegraphics[width=25mm]{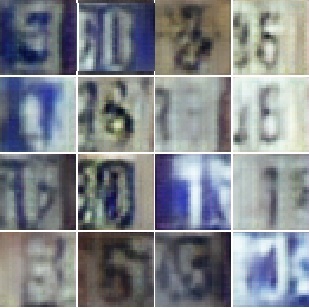} \\ 
		(d) CNN & (e) FCC-GAN-S & (f) FCC-GAN-P \\
		\includegraphics[width=25mm]{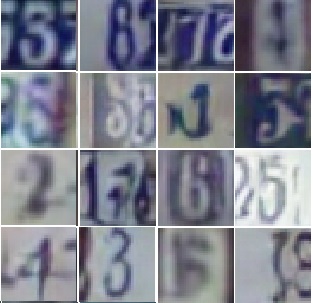} & 
			\includegraphics[width=25mm]{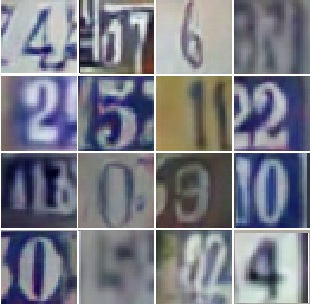} & 
				\includegraphics[width=25mm]{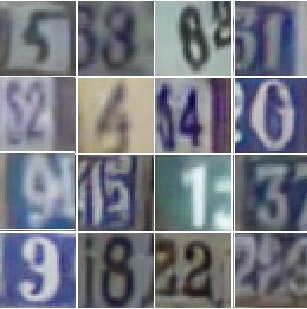} \\ 
		(g) CNN & (h) FCC-GAN-S & (i) FCC-GAN-P \\
\end{tabular}
\caption{$32\times32$ pixel SVHN images for standard GAN training. Images generated by conventional CNN and proposed FCC-GAN models after 1 epoch (a-c), 
10 epochs (d-f), and 100 epochs (g-i).}
\label{fig:dcgansvhnoutput}
\end{figure}

\begin{table}[ht!]
\small
  \begin{center}
    \caption{Inception scores (higher is better) for standard GAN training for conventional CNN and proposed FCC-GAN architectures.}
    \label{tab:inceptionscoresgantraining}
    \begin{tabular}{c|c|c|c}
			\hline
			\textbf{Model}&\textbf{MNIST}&\textbf{CIFAR-10}&\textbf{SVHN} \\ \hline
            CNN&8.68 $\pm$ 0.03&6.37 $\pm$ 0.07&2.97 $\pm$ 0.03 \\ \hline
            FCC-GAN-S&9.34 $\pm$ 0.02&6.72 $\pm$ 0.08&3.08 $\pm$ 0.04 \\ \hline
            FCC-GAN-P&\textbf{9.48 $\pm$ 0.03}&\textbf{7.25 $\pm$ 0.05}&\textbf{3.23 $\pm$ 0.03} \\ \hline
    \end{tabular}
  \end{center}
\end{table}

\begin{table}[ht!]
\small
  \begin{center}
    \caption{FIDs (lower is better) for standard GAN training for conventional CNN and proposed FCC-GAN architectures.}
    \label{tab:fidgantraining}
    \begin{tabular}{c|c|c|c}
			\hline
			\textbf{Model}&\textbf{MNIST}&\textbf{CIFAR-10}&\textbf{SVHN} \\ \hline
            CNN&8.07 $\pm$ 0.20&40.84 $\pm$ 0.34&19.22 $\pm$ 0.16 \\ \hline
            FCC-GAN-S&4.69 $\pm$ 0.12&41.47 $\pm$ 0.17&18.46 $\pm$ 0.11 \\ \hline
            FCC-GAN-P&\textbf{4.23 $\pm$ 0.02}&\textbf{34.07 $\pm$ 0.18}&\textbf{15.56 $\pm$ 0.11} \\ \hline

    \end{tabular}
  \end{center}
\end{table}

Figure~\ref{fig:dcgancifar10output} shows the images generated by the three models for the CIFAR-10 dataset. Comparing the images generated after epoch 1 and 35, we observe that the FCC-GAN models' outputs are visually superior to those of the CNN model. The results on SVHN show similar trends to the experiments for MNIST and CIFAR-10 (see Figure~\ref{fig:dcgansvhnoutput} for the SVHN images produced by the three models for epochs 1, 10, and 100).

In Table~\ref{tab:inceptionscoresgantraining} and Table~\ref{tab:fidgantraining}, we report the best Inception scores and FIDs obtained by the three models. FCC-GAN models obtain better scores than the CNN model. Specifically, FCC-GAN-P model obtains the best results and outperforms the other models in terms of both Inception score and FID. 
\begin{figure}[ht!]
	\centering
	\small
	\begin{tabular}{ccc}
		\includegraphics[width=25mm]{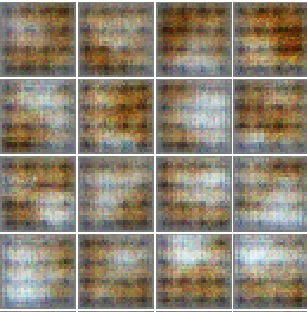} & 
			\includegraphics[width=25mm]{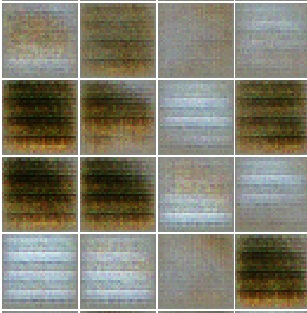} & 
				\includegraphics[width=25mm]{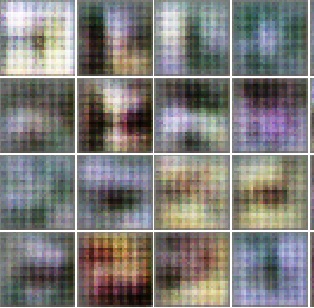}\\ 
		(a) CNN & (b) FCC-GAN-S & (c) FCC-GAN-P \\
		\includegraphics[width=25mm]{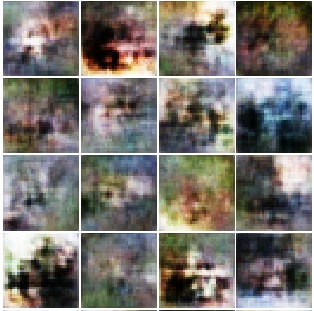}& 
			\includegraphics[width=25mm]{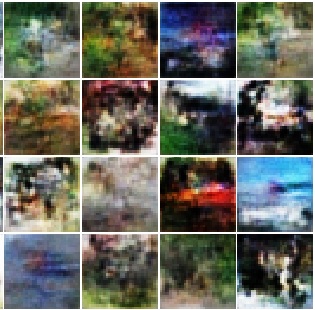} & 
				\includegraphics[width=25mm]{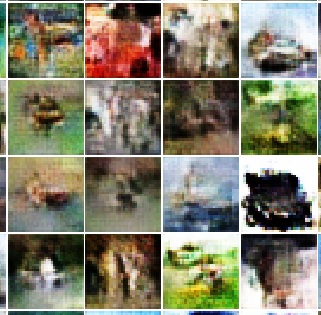} \\
		(d) CNN & (e) FCC-GAN-S & (f) FCC-GAN-P \\
		\includegraphics[width=25mm]{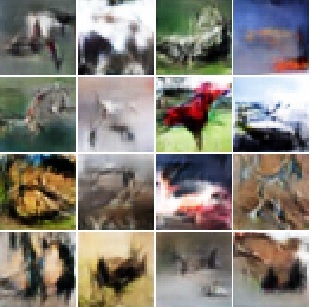} & 
			\includegraphics[width=25mm]{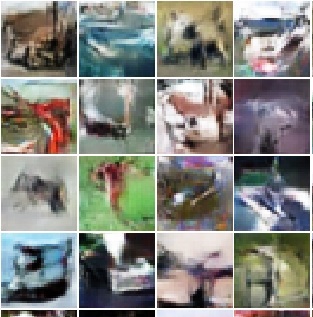} & 
				\includegraphics[width=25mm]{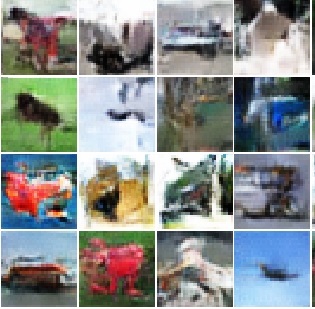} \\ 
		(g) CNN & (h) FCC-GAN-S & (i) FCC-GAN-P \\
\end{tabular}
\caption{$32\times32$ pixel CIFAR-10 images for WGAN training. Images generated by conventional CNN and proposed FCC-GAN models after 1 epoch (a-c), 10 epochs (d-f), and 150 epochs (g-i)}
\label{fig:wgancifar10output}
\end{figure}
\begin{table}[ht!]
\small
  \begin{center}
    \caption{Inception scores (higher is better) for WGAN training for conventional CNN and proposed FCC-GAN models.}
    \label{tab:inceptionscoreswgantraining}
    \begin{tabular}{c|c|c|c}
			\hline
			\textbf{Model}&\textbf{MNIST}&\textbf{CIFAR-10}&\textbf{SVHN} \\ \hline
            CNN&7.00 $\pm$ 0.03&5.33 $\pm$ 0.06&2.83 $\pm$ 0.02 \\ \hline
            FCC-GAN-S&7.25 $\pm$ 0.04&5.42 $\pm$ 0.06&\textbf{3.03 $\pm$ 0.02} \\ \hline
            FCC-GAN-P&\textbf{8.67 $\pm$ 0.03}&\textbf{5.60 $\pm$ 0.07}&3.00 $\pm$ 0.01 \\ \hline
    \end{tabular}
  \end{center}
\end{table}

\begin{table}[ht!]
\small
  \begin{center}
    \caption{FIDs (lower is better) for WGAN training for conventional CNN and proposed FCC-GAN models.}
    \label{tab:fidwgantraining}
    \begin{tabular}{c|c|c|c}
			\hline
			\textbf{Model}&\textbf{MNIST}&\textbf{CIFAR-10}&\textbf{SVHN} \\ \hline
            CNN&30.17 $\pm$ 0.25&62.21 $\pm$ 0.20&53.60 $\pm$ 0.12 \\ \hline
            FCC-GAN-S&30.76 $\pm$ 0.01&58.33 $\pm$ 0.15&\textbf{43.82 $\pm$ 0.48} \\ \hline
            FCC-GAN-P&\textbf{20.26 $\pm$ 0.18}&\textbf{56.89 $\pm$ 0.22}&69.60 $\pm$ 0.30 \\ \hline
    \end{tabular}
  \end{center}
\end{table}
\textbf{Results on WGAN Training:} 
Results for WGAN training are similar to those for standard GAN. Tables~\ref{tab:inceptionscoreswgantraining} and ~\ref{tab:fidwgantraining} show the best Inception scores and FIDs obtained by the three models. We see that for all experiments, FCC-GAN models achieved better scores than the CNN model in WGAN training as well. In particular, FCC-GAN-P model was superior among the three models outperforming the other two in two of the three datasets. For space constraints, we only show the generated images for CIFAR-10 in Figure~\ref{fig:wgancifar10output}. We observe that the images generated by the FCC-GAN models are visually superior to those of the CNN models.

\begin{figure}[ht!]
\centering
\begin{tabular}{cc}
	\includegraphics[width=39mm]{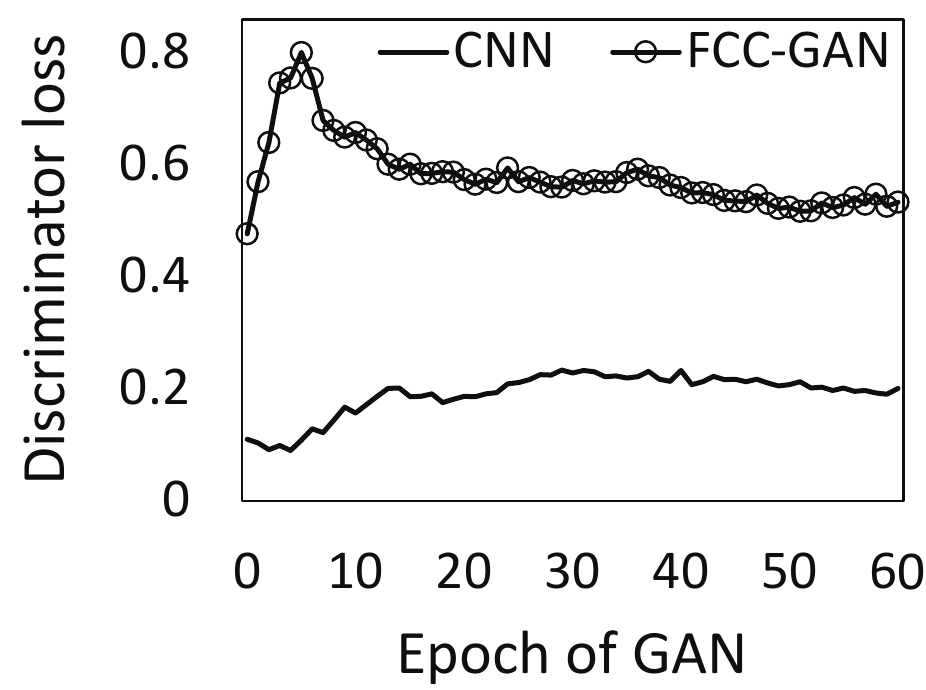} &
	    	\includegraphics[width=39mm]{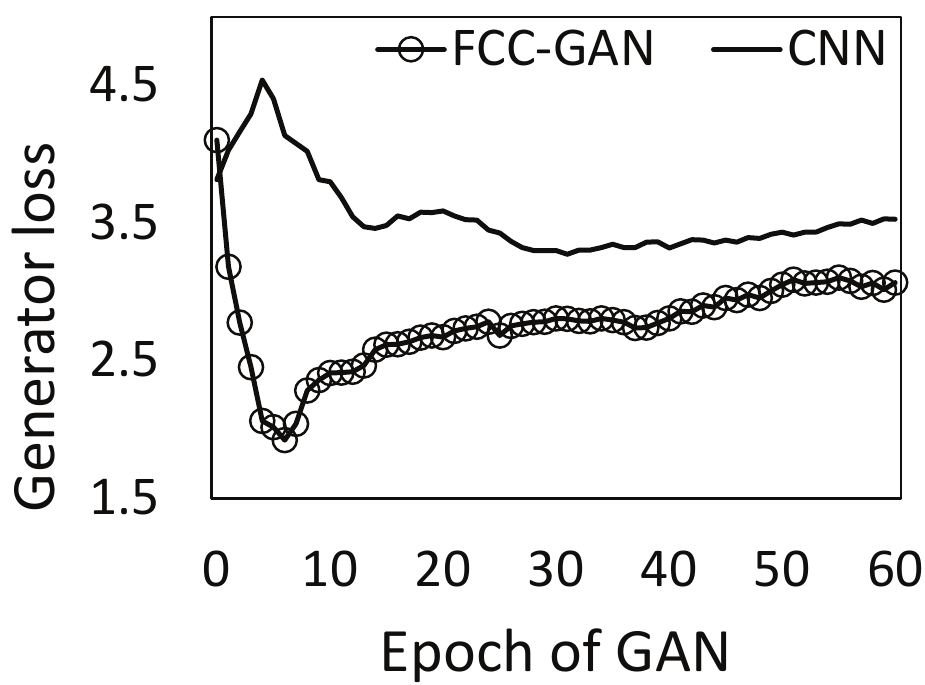} \\
	(a) & (b) \\
\end{tabular}
\caption{(a) Discriminator loss of the CNN and FCC-GAN models for MNIST dataset (b) Generator loss of the CNN and FCC-GAN models for MNIST dataset}
\label{fig:gendislossvaryepoch}
\end{figure}

\textbf{Higher Discriminator Loss:}
In this section, we empirically demonstrate that the FCC-GAN discriminator has higher loss values than the CNN discriminator.
Consider Figure~\ref{fig:gendislossvaryepoch}(a) where we plot the loss of CNN and FCC-GAN (with pooling) models over different epochs of standard GAN training on MNIST. We observe that the discriminator of the FCC-GAN model has a higher loss values than that of the CNN model across all epochs. On average, FCC-GAN model has 62\% higher loss than the CNN model. This higher loss has the potential to provide larger gradients to help the generator learn the data distribution faster.

\textbf{Lower Generator Loss:} 
We compare the loss of the generators of the two architectures in Figure~\ref{fig:gendislossvaryepoch}(b) for the same experiment on MNIST. The FCC-GAN  generator achieves lower loss values compared to the CNN model. In particular, comparing the loss pattern of the two generators at the early stages of training (epochs 1-5), we see that the loss of the FCC-GAN model sharply decreases in contrast to an increase of loss for the CNN model. As training progresses, the discriminator improves and the generator loss starts to increase (after epoch 7 in the figure).

\begin{table}[ht!]
\begin{center}
\small
\caption{Number of epochs needed by the CNN and FCC-GAN models to achieve a desired Inception score for MNIST dataset. FCC-GAN model converges much faster than the CNN model.}
\label{tab:inceptionmnistepoch}
\begin{tabular}{c|c|c|c|c|c}
\hline
&\multicolumn{5}{c}{Inception score (higher is better)}\\ \hline
Method & $\geq6.00$&$\geq7.00$&$\geq8.0$ &$\geq8.5$ & $\geq9.0$\\ \hline \hline
FCC-GAN&1&2&3&4&8\\ \hline
CNN&3&5&12&26&-\\ \hline
\end{tabular}
\end{center}
\end{table}

\textbf{Faster Convergence:}
The loss pattern of the generator and discriminator of FCC-GAN is actually associated with a faster learning of the data distribution. To illustrate this, in Table~\ref{tab:inceptionmnistepoch}, we report the number of epochs needed by both models to achieve a desired Inception score for a standard GAN training on MNIST. The Inception score for the MNIST real dataset is 9.95. From Table~\ref{tab:inceptionmnistepoch}, we see that, FCC-GAN obtained an Inception score above 6.00 after just 1 epoch of training, whilst CNN requires 3 epochs. After 3 epochs, FCC-GAN achieves an score above 8.0 compared to 11 epochs needed by CNN. Notably, after just 8 epochs, Inception score of the FCC-GAN reached above 9.0. The CNN model did not achieve an Inception score above 9.0 even after running the model for 100 epochs (above 187K generator iterations). Table~\ref{tab:inceptioncifar10epoch} provides similar comparisons for a standard GAN training on CIFAR-10. We see that FCC-GAN achieves Inception scores above 6.5 after epoch 41 and 7.0 after epoch 112. The CNN model could not obtain an Inception score above 6.5 even after running for 150 epochs (above 234K generator iterations). These results demonstrate a huge speedup in convergence of FCC-GAN compared to the standard CNN architecture.
\begin{table}[ht!]
\begin{center}
\small
\caption{Number of epochs needed by the CNN and FCC-GAN models to achieve a desired Inception score for CIFAR-10 dataset. FCC-GAN model converges much faster than the CNN model.}
\label{tab:inceptioncifar10epoch}
\begin{tabular}{c|c|c|c|c|c}
\hline
&\multicolumn{5}{c}{Inception score (higher is better)}\\ \hline
Method & $\geq5.0$&$\geq5.5$&$\geq6.0$ &$\geq6.5$ & $\geq7.0$\\ \hline\hline
FCC-GAN&18&24&32&41&112\\ \hline
CNN&26&34&81&-&-\\ \hline
\end{tabular}
\end{center}
\end{table}
\begin{figure}[ht!]
	\centering
	\small
	\begin{tabular}{ccc}
		\includegraphics[width=25mm]{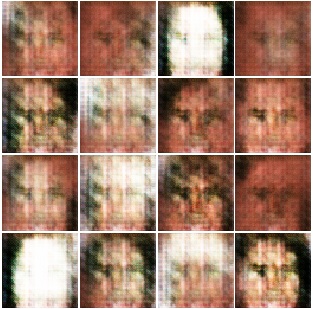} & 
		    \includegraphics[width=25mm]{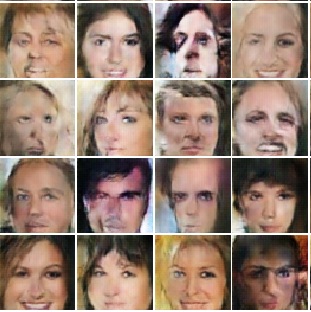} &
		    	\includegraphics[width=25mm]{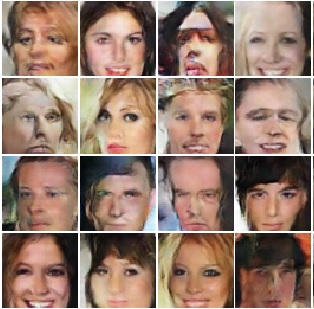} \\
		(a) 1 epoch & (b) 5 epochs & (c) 10 epochs \\
		\includegraphics[width=25mm]{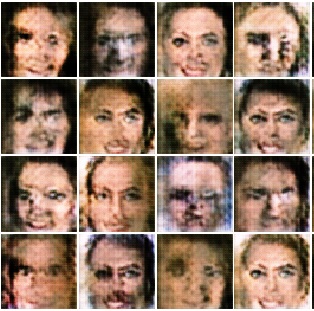} &
    		\includegraphics[width=25mm]{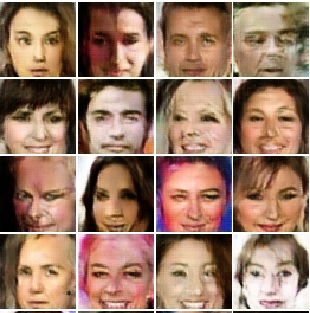} &
    			\includegraphics[width=25mm]{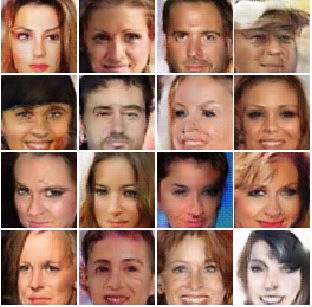} \\ 
		(d) 1 epoch & (e) 5 epochs & (f) 10 epochs \\
\end{tabular}
\caption{$64\times64$ pixel CelebA images generated by conventional CNN model (a-c) and proposed FCC-GAN model (d-f) after different epochs of an standard GAN training}
\label{fig:stdgancelebaoutput}
\end{figure}

\textbf{Experiments on $64\times64$ pixel CelebA dataset:} 
To assess the performance of FCC-GAN on higher resolution datasets, we evaluated it on
$64\times64$ resolution CelebA~\citep{celeba} daataset. We trained both CNN and FCC-GAN models for 50 epochs using standard GAN training. The models were similar to previous experiments with an additional convolution layer for upsampling (downsampling) in the generator (discriminator) to match with the higher resolution. Detailed architecture can be found in Appendix~\ref{sec:celebanetwork}. Adam optimizer was used with a learning rate of 0.0001 and decay of 0.0001. Figure~\ref{fig:stdgancelebaoutput} shows the generated images by both CNN and FCC-GAN models after different epochs of the training. The higher image quality and faster learning of the FCC-GAN model are clearly visible in the figure for all epochs.


\begin{table}
  \begin{center}
    \caption{Inception scores (higher is better) for CNN and FCC-GAN models on CIFAR10 dataset over different architectures of varying convolution layers. $N_1$ and $N_2$ denote the number of convolution layers used in the generator and discriminator of CNN model, respectively. Note that for FCC-GAN model, the first (last) convolution layer in the generator (discriminator) is removed to accommodate three fully connected layers. The models were trained for 50 epochs using standard GAN training.}
    \label{tab:inceptionvaryconvlayers}
    \begin{tabular}{C{1.5cm}|C{1cm}|C{1cm}|C{1cm}|C{1cm}}
			\hline
			&\multicolumn{4}{c}{Number of convolution layers, $N_1+N_2$} \\ \hline
			Method&4+4&4+7&7+4&7+7 \\ \hline
            CNN&5.98&5.63&3.69&4.24 \\ \hline
            FCC-GAN&6.69&6.91&6.84&6.84 \\ \hline
    \end{tabular}
  \end{center}
\end{table}

\textbf{Experiments with ResNet:}
In this section, we demonstrate the performance of FCC-GAN architecture with ResNet models. Previously, GAN models~\cite{improvedwgan,sngan} based on ResNet architecture showed significant performance improvement over models build with simple convolution layers. To verify whether FCC-GAN architecture can benefit ResNet GAN models, we modify the ResNet model used by the WGAN-GP method for CIFAR10 dataset, and add fully connected layers at the beginning (end) of the generator (discriminator) network. We added three FC layers in the generator having 64, 512, and 2048 nodes, respectively, in each layer. The FC configuration is similar to the one shown in Table~\ref{tab:hybridmodelcifar10} except that the last (third) FC layer has 2048 nodes to match with the input filter of the subsequent ResNet blocks (as used by WGAN-GP experiments). We added two FC layers having 16 and 1 node at the end of the ResNet discriminator. The ResNet blocks in the original WGAP-GP discriminator outputs a vector of 128 dimensions, thus we only needed two FC layers to reduce the dimension further to 16 and then produce the final output. The value of all hyper-parameters, e.g., batch size, optimization algorithm, learning rate, and decay, were kept similar to the ones used in the original experiments~\cite{improvedwgan}. The ResNet GAN model was run for $100K$ generator iterations on the CIFAR10 dataset. We found the standard GAN loss function to perform better than the WGAN loss function for FCC-GAN ResNet models. Table~\ref{tab:inceptionresnet} shows the best Inception scores obtained on the CIFAR10 dataset by the original WGAN-GP ResNet model and FCC-GAN ResNet model in our unsupervised GAN training. Our FCC-GAN architecture on ResNet achieved higher (better) Inception scores than the original ResNet model used by WGAN-GP experiments. The results demonstrate that the proposed FCC-GAN architecture can substantially improve the performance of ResNet GAN models.
\begin{table}[ht!]
\small
  \begin{center}
    \caption{Inception scores obtained by the original WGAN-GP ResNet model and FCC-GAN ResNet model in unsupervised GAN training on CIFAR10 dataset.}
    \label{tab:inceptionresnet}
    \begin{tabular}{c|c}
			\hline
			\textbf{Model} & \textbf{Inception Score} \\ \hline
            WGAN-GP ResNet~\cite{improvedwgan} & 7.86 $\pm$ 0.07 \\ \hline
            FCC-GAN ResNet & \textbf{8.02 $\pm$ 0.08} \\ \hline
    \end{tabular}
  \end{center}
\end{table}

\textbf{Impact of convolution network depth on CNN and FCC-GAN architecture:}
One may argue that the performance benefit of FCC-GAN model results from higher network depths (achieved due to the inclusion of FC layers in both generator and discriminator). To evaluate this argument, we perform GAN experiments by increasing the network depth of CNN models and compare the results with those of FCC-GAN models. We increase the number of convolution layers of CNN models from 4 (as shown in Table~\ref{tab:hybridmodelcifar10}) to 7 in the generator network, discriminator network, and both networks together. We achieved this by adding an unit-stride convolution layer with $3\times3$ kernel after (before) each upsampling (downsampling) convolution layer in the generator (discriminator) (Please see Appendix~\ref{sec:architectureconvdepthexp} for detailed network architecture). Then we trained the models on CIFAR10 dataset using standard GAN training for 50 epochs. The best Inception scores are shown in the third row of Table~\ref{tab:inceptionvaryconvlayers}. We see that increasing the depth of CNN model did not result in better Inception scores. In fact, CNN models with 7 convolution layers in the generator did not converge (obtained low Inception scores of 3.69 and 4.64). However, similar increase of depth for FCC-GAN models results in a performance boost. As can be seen from the Inception scores of last row of Table~\ref{tab:inceptionvaryconvlayers}, larger architectures (4+7, 7+4, and 7+7) get better Inception scores than the smallest (4+4) architecture for FCC-GAN models. We also observe that the smallest FCC-GAN model (4+4) obtains better Inception score than all four CNN models. These results imply that the benefits of FCC-GAN architecture results necessarily from the inclusion of FC layers, and not merely from the increase of network depth. Choosing a different set of hyper-parameters might have an impact on the performance of the higher depth CNN models; however, for fair comparison we stick to the same hyper-parameters for all experiments. Although increasing convolution network depth did not improve performance of the simple CNN models (as used in our experiments), high depth networks are usually effective for complex architectures such as ResNet~\citep{improvedwgan}, particularly in generating high resolution images~\citep{acgan,sngan,largescalegan}.

\begin{figure}[ht!]
\centering
\begin{tabular}{cc}
		\includegraphics[width=39mm]{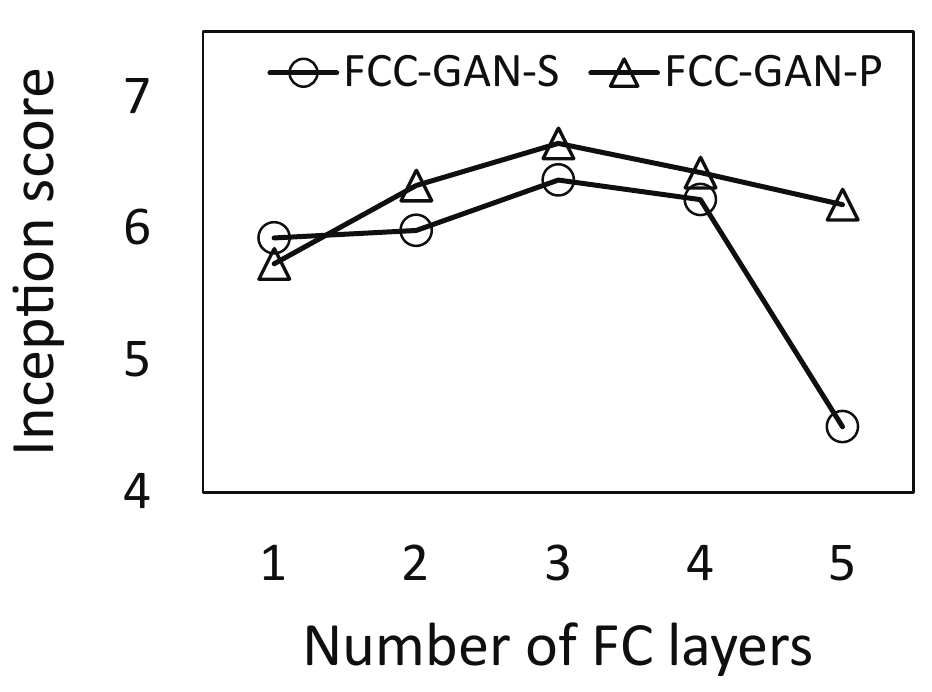} &
		    \includegraphics[width=39mm]{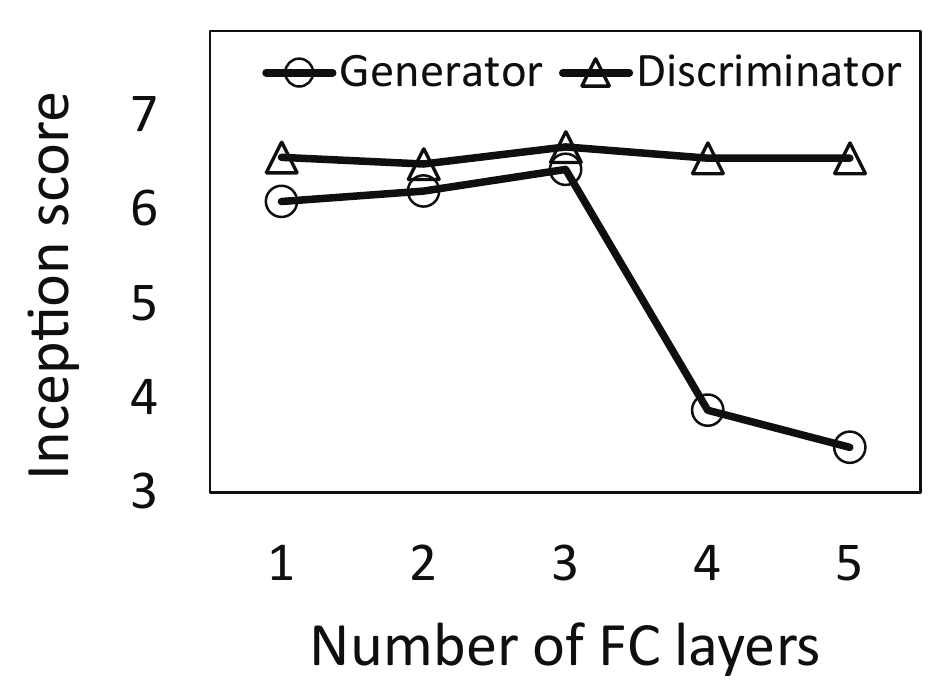} \\
		(a) & (b) \\
\end{tabular}
\caption{(a) Inception score for varying the number of FC layers in both the generator and discriminator simultaneously (b) Inception scores for varying the number of FC layers in a single network (e.g., generator) keeping the other network (e.g., discriminator) fixed to a CNN model. The results reported are for FCC-GAN model with pooling.}
\label{fig:inceptionvaryfc}
\end{figure}

\textbf{Impact of FC network depth on FCC-GAN architecture:}
It may be interesting to examine how the performance of the FCC-GAN model changes when the number of layers of the FC network is varied. We conduct two different studies to examine the impact of the number of FC layers in the FCC-GAN architecture. In one,  we keep the number of FC layers in both the generator and discriminator same, and vary them from one to five layers. In the other, we vary the number of FC layers in one network, i.e., generator (discriminator), while we use a fixed CNN model for the other network, i.e., discriminator (generator). For both studies, we train the FCC-GAN models on CIFAR-10 dataset using standard GAN training for up to 50 epochs. Figure~\ref{fig:inceptionvaryfc} shows the plot of Inception scores obtained over number of FC layers. As can be seen from Figure~\ref{fig:inceptionvaryfc}(a), the best score for CIFAR-10 is obtained when the network has three FC layers in the generator and discriminator. Increasing the number of layers beyond four decreases the scores for FCC-GAN-S drastically, though FCC-GAN-P still achieves moderately good scores. It shows FCC-GAN-P is a more robust architecture than FCC-GAN-S in terms of such variations. For the other study, we find from Figure~\ref{fig:inceptionvaryfc}(b) that, the performance of FCC-GAN degrades significantly for four or more FC layers in the generator. However, the performance remains very stable (good) for such variations in the discriminator. Interestingly, for both studies, the best Inception score is obtained at three FC layers.

\subsection{Stability of FCC-GAN Architecture}
\begin{figure}[ht!]
	\centering
	\small
	\begin{tabular}{ccc}
		\includegraphics[width=25mm]{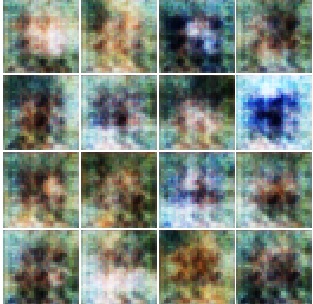} & 
			\includegraphics[width=25mm]{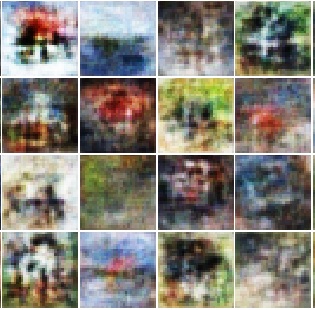} & 
				\includegraphics[width=25mm]{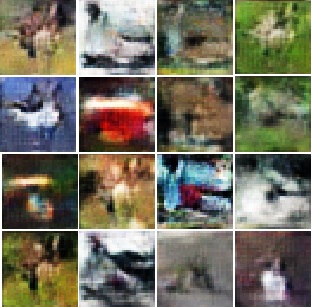}\\ 
		(a) CNN & (b) FCC-GAN-S & (c) FCC-GAN-P \\
		\includegraphics[width=25mm]{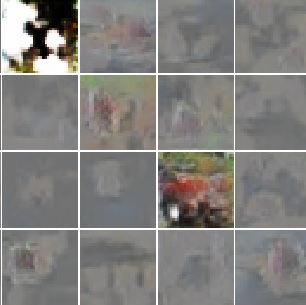} & 
			\includegraphics[width=25mm]{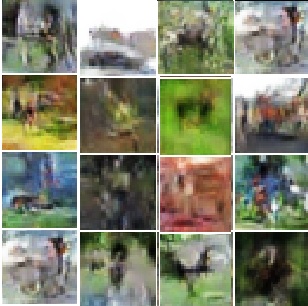} & 
				\includegraphics[width=25mm]{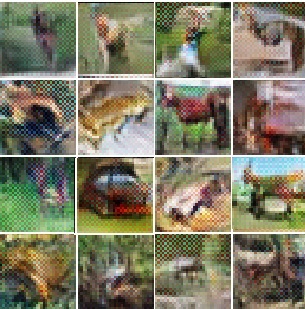} \\
		(d) CNN & (e) FCC-GAN-S & (f) FCC-GAN-P \\
		\includegraphics[width=25mm]{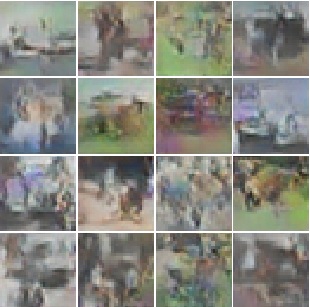} & 
			\includegraphics[width=25mm]{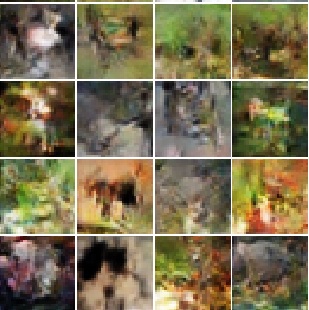} & 
				\includegraphics[width=25mm]{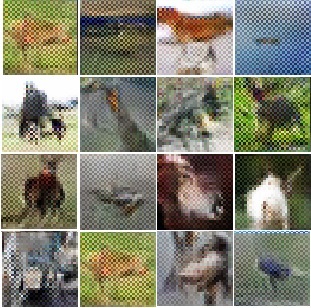} \\ 
		(g) CNN & (h) FCC-GAN-S & (i) FCC-GAN-P \\
    \end{tabular}
    \caption{Images generated for (a-c) SGD optimization algorithm, (d-f) No BN in discriminator, and (g-i) No BN in both the generator and discriminator}
    \label{fig:statibilityoutput}
\end{figure}

We examined the stability of the proposed FCC-GAN models with respect to a variety of experimental settings. First, we examined the stability with respect to the presence and absence of batch-normalization (BN) layers in the generator and discriminator, following the similar experiments by~\citep{wgan}. Then, we varied the optimization algorithm. We ran experiments on CIFAR-10 dataset for up to 50 epochs to verify stability of the models. Table~\ref{tab:stabilityresults} reports the Inceptions scores and Figure~\ref{fig:statibilityoutput} shows some comparisons of generated samples. The results are descried below.

\begin{table}[h!]
	\small
	\setlength\tabcolsep{5.0pt}
  \begin{center}
    \caption{Inception scores on CIFAR-10 dataset for stability experiments. Standard deviations are omitted. Very low Inception scores (* marked) indicate the corresponding model completely failed to produce any recognizable images.}
    \label{tab:stabilityresults}
    \begin{tabular}{c|c|c|c}
    \hline
        Method&CNN&FCC-GAN-S&FCC-GAN-P\\\hline
        RMSProp&5.9512224&5.863833&\textbf{6.860175}\\ \hline
        SGD&1.6547798*&2.4576838*&\textbf{3.8250222}\\ \hline
        NBN-BN&5.717898&6.0637116&\textbf{6.1960044}\\ \hline
        BN-NBN&2.213803*&3.8931298&\textbf{4.8963203}\\ \hline
        NBN-NBN&3.4395053*&3.522677&\textbf{4.0727606}\\ \hline
    \end{tabular}
  \end{center}
\end{table}

\textbf{Impact of BN layers.} For this study, we trained the models after sequentially removing BN from the generator (NBN-BN), discriminator (BN-NBN), and from both the generator and discriminator (NBN-NBN). 

\begin{itemize}
\item No BN in the generator: When we removed BN from the generator only, all models obtained good quality images and inception scores. However, the scores obtained by the FCC-GAN models were better than those of CNN model. Another point to note is that the scores found after removing BN from the generator was worse than that of the generator with BN. This implies that the use of BN in the generator has a substantial impact in improving performance of GAN models both in terms image quality and inception scores.

\item No BN in the discriminator: Wen we removed BN from the discriminator, the CNN model failed to produce any recognizable outputs (see figure~\ref{fig:statibilityoutput}(d)). Our FCC-GAN model also failed to produce very good quality images. However, FCC-GAN-P model was still able to produce some recognizable images (see figure~\ref{fig:statibilityoutput}(f)). The images were worse than those of BN-inclusive discriminator. 

\item No BN in the generator and discriminator: The results obtained after removing BN from both the generator and discriminator was similar to the results obtained after removing BN from the discriminator only. The CNN model failed to capture the distribution while our FCC-GAN models, specifically the FCC-GAN-P was able to produce some recognizable images (see Figure~\ref{fig:statibilityoutput}(i)). 
\end{itemize}

In summary, for all cases, FCC-GAN models obtained better image quality and Inception scores (see Table~\ref{tab:stabilityresults}). In particular, FCC-GAN-P outperformed others.

\textbf{Impact of optimization algorithm.} 
We changed the default optimization algorithm from ADAM to SGD and RMSProp with a learning rate of 0.0001 and a decay of 0.00001. Table~\ref{tab:stabilityresults} shows the Inception scores obtained after 50 epochs of the training for the three optimization algorithms. For RMSProp, all three models are stable and obtain good Inception scores. However, our FCC-GAN models achieve better scores than the CNN model, and FCC-GAN-P outperforms the others. For SGD, the CNN model fails to produce any recognizable images after 50 epochs (Figure~\ref{fig:statibilityoutput}(a)). FCC-GAN-S seems to have captured some properties of the data, and produces comparably better images than CNN (Figure~\ref{fig:statibilityoutput}(b)). In contrast, FCC-GAN-P generates very good quality images (Figure~\ref{fig:statibilityoutput}(c)).

\textbf{Summary of results:}
We verified the effectiveness of FCC-GAN architecture on four benchmark image datasets. We also verified the applicability of the architecture with respect to two different objective functions commonly used in GAN training algorithms. In all experiments, FCC-GAN architecture generated higher quality images and obtained better Inception scores and FIDs than the conventional CNN architecture used be existing GAN models. In particular, FCC-GAN converged much faster than the CNN architecture. The training stability of the proposed architecture was also verified across a variety of experiment settings, which shows a higher stability of FCC-GAN compared to CNN. 

\section{Conclusion}
In this paper, we proposed FCC-GAN architecture for GANs. In our architecture, the generator and discriminator networks consist of deep fully connected and convolution layers, in contrast to conventional deep convolution networks. We performed experiments on four benchmark image datasets and showed that our proposed architecture generates higher quality samples, obtains better Inception scores and Fr\'etchet Inception Distances, and converges faster than the conventional architecture. We also empirically validated its stability in a wide variety of experimental settings.

One key advantage is that the proposed architecture can be applied in combination with any other GAN settings. For example, in this study, all of our experiments were limited to unsupervised training (without using labels). Thus, a straightforward extension of the work is to study the effectiveness of the FCC-GAN architecture in semi-supervised and conditional GAN settings. Another possible extension is to adapt FCC-GAN architecture for more complex networks such as ResNet.

\bibliographystyle{ACM-Reference-Format}
\bibliography{fccgan}

\newpage

\appendix

\section{Details of network architecture and experimental settings}
\label{sec:modelarchdetails}
\subsection{Model architecture used for CIFAR-10 and SVHN} 
Table~\ref{tab:hybridmodelcifar10v2} shows the FCC-GAN model architecture for CIFAR-10 and SVHN dataset. The images have the dimension $32 \times 32 \times 3$. In the generator, we used three FC layers that map the 100-dimensional noise to 4096-dimensional intermediate features. The linear features are then reshaped to a spatial extent of dimension $4 \times 4 \times 256$. Then four transposed convolution layers map these features to output images of dimension $32 \times 32 \times 3$. ReLU activation is used for all layers except the output which uses Tanh activation. Batch-normalization (BN) is used after the last FC layer and all the convolution layers except the final output. In the discriminator, three convolution layers extract image features of dimension $4 \times 4 \times 256$. The features are then flattened to a 4096-dimensional feature vector. These features are then mapped to a low-dimensional space by four FC layers before the final output. LeakyReLU (LReLU) with a slope of 0.2 is used as the activation function for all layers except the final output. Sigmoid is used as the output activation for standard GAN~\citep{gan} training while no activation is used for Wasserstein GAN~\citep{wgan}. For FCC-GAN-P model, we use average pooling instead of strided-convolution for downsampling. This implies we replace each CONV(x,y,$2$) layer in the discriminator with an equivalent CONV(x,y,$1$) layer followed by an average pooling with a pool size of $2\times2$ for downsampling. Table~\ref{tab:cnnmodelcifar10} shows the conventional CNN model used for CIFAR-10 and SVHN. Note that the first (last) convolution layer in the generator (discriminator) of CNN models can be considered a fully connected layer. 
\begin{table}[ht!]
\small
  \begin{center}
    \caption{FCC-GAN architecture used for CIFAR-10 and SVHN}
    \label{tab:hybridmodelcifar10v2}
    \begin{tabular}{c|c}
        \multicolumn{2}{c}{} \\ \hline 
        Generator&Discriminator\\ \hline
        Input: Z(100)&Input: Image (32,32,3)\\
        FC(64), ReLU&CONV(64,4,2), BN, LReLU\\
        FC(512), ReLU&CONV(128,4,2), BN, LReLU\\
        FC(4096), BN&CONV(256,4,2), BN, LReLU\\
        Reshape (4,4,256)&Flatten (4096)\\
        CONVT(128,4,2), BN, ReLU&FC(512), LReLU\\
        CONVT(64,4,2), BN, ReLU&FC(64), LReLU\\
        CONVT(3,4,3), Tanh&FC(16), LReLU\\
        Output: (32, 32,3)&FC(1), Sigmoid\\
        &Output: 1\\
    \end{tabular}
  \end{center}
\end{table}

\begin{table}[ht!]
\small
  \begin{center}
    \caption{CNN architecture used for CIFAR-10 and SVHN}
    \label{tab:cnnmodelcifar10}
    \begin{tabular}{c|c}
        \multicolumn{2}{c}{} \\ \hline 
        Generator&Discriminator\\ \hline
        Input: Z(100)&Input: (32,32,3)\\
        R(1,1,100)&CONV(64,4,2), BN, LReLU\\
        CONVT(256,4,1), BN, ReLU&CONV(128,4,2), BN, LReLU\\
        CONVT(128,4,2), BN, ReLU&CONV(256,4,2), BN, LReLU\\
        CONVT(64,4,2), BN, ReLU&CONV(1,4,1), Sigmoid\\
        CONVT(3,4,3), Tanh&Output: 1\\
        Output: (32, 32, 3)&\\
    \end{tabular}
  \end{center}
\end{table}

\subsection{Model architecture used for MNIST}
\label{sec:mnistnetarchitecture}
MNIST image size is $28 \times 28 \times 1$. Thus, the convolution layers used for MNIST have different kernel and filter sizes than those used for CIFAR-10 and SVHN. Table~\ref{tab:hybridmodelmnist} shows the FCC-GAN architecture while Table~\ref{tab:cnnmodelmnist} shows the CNN architecture. 

\begin{table}[ht!]
\small
  \begin{center}
    \caption{FCC-GAN architecture used for MNIST}
    \label{tab:hybridmodelmnist}
    \begin{tabular}{c|c}
        \multicolumn{2}{c}{} \\ \hline 
        Generator&Discriminator\\ \hline
        Input: Z(100)&Input: (28,28,1)\\
        FC(64), ReLU&CONV(32,3,2), BN, LReLU\\
        FC(512), ReLU&CONV(64,3,2), BN, LReLU\\
        FC(1152), BN&CONV(128,3,2), BN, LReLU\\
        Reshape (3,3,128)&Flatten (1152)\\
        CONVT(64,3,2), BN, ReLU&FC(512), LReLU\\
        CONVT(32,3,2), BN, ReLU&FC(64), LReLU\\
        CONVT(1,3,2), Tanh&FC(16), LReLU\\
        Output: (28, 28, 1)&FC(1), Sigmoid\\
        &Output: 1\\
        
    \end{tabular}
  \end{center}
\end{table}

\begin{table}[ht!]
\small
  \begin{center}
    \caption{CNN architecture used for MNIST}
    \label{tab:cnnmodelmnist}
    \begin{tabular}{c|c}
        \multicolumn{2}{c}{} \\ \hline 
        Generator&Discriminator\\ \hline
        Input: Z(100)&Input: (28,28,1)\\
        R(1,1,100)&CONV(32,3,2), BN, LReLU\\
        CONVT(128,3,1), BN, ReLU&CONV(64,3,2), BN, LReLU\\
        CONVT(64,3,2), BN, ReLU&CONV(128,3,2), BN, LReLU\\
        CONVT(32,3,2), BN, ReLU&CONV(1,3,1), Sigmoid\\
        CONVT(1,3,2), Tanh&Output: 1\\
    \end{tabular}
  \end{center}
\end{table}

\subsection{Model architecture used for CelebA}
\label{sec:celebanetwork}
Table ~\ref{tab:fccganarchceleba} and ~\ref{tab:cnnarchceleba} show the network architectures used for $64\times64$ pixel CelebA dataset. Note that the network configuration is similar to those of CIFAR10 dataset except an additional convolution layer in both the generator and discriminator. 

\begin{table}[h!]
\small
  \begin{center}
    \caption{FCC-GAN architecture used for CelebA}
    \label{tab:fccganarchceleba}
    \begin{tabular}{c|c}
        \multicolumn{2}{c}{} \\ \hline 
        Generator&Discriminator\\ \hline
        Input: Z(100)&Input: Image (64,64,3)\\
        FC(64), ReLU&CONV(64,4,2), BN, LReLU\\
        FC(512), ReLU&CONV(128,4,2), BN, LReLU\\
        FC(8192), BN&CONV(256,4,2), BN, LReLU\\
        Reshape (4,4,512), BN&CONV(512,4,2), BN, LReLU\\
        CONVT(256,4,2)&Flatten (8192)\\
        CONVT(128,4,2), BN, ReLU&FC(512), LReLU\\
        CONVT(64,4,2), BN, ReLU&FC(64), LReLU\\
        CONVT(3,4,3), Tanh&FC(16), LReLU\\
        Output: (32, 32,3)&FC(1), Sigmoid\\
        &Output: 1\\
    \end{tabular}
  \end{center}
\end{table}

\begin{table}[h!]
\small
  \begin{center}
    \caption{CNN architecture used for CelebA}
    \label{tab:cnnarchceleba}
    \begin{tabular}{c|c}
        \multicolumn{2}{c}{} \\ \hline 
        Generator&Discriminator\\ \hline
        Input: Z(100)&Input: (64,64,3)\\
        R(1,1,100)&CONV(64,4,2), BN, LReLU\\
        CONVT(512,4,1), BN, ReLU&CONV(128,4,2), BN, LReLU\\
        CONVT(256,4,1), BN, ReLU&CONV(256,4,2), BN, LReLU\\
        CONVT(128,4,2), BN, ReLU&CONV(512,4,2), BN, LReLU\\
        CONVT(64,4,2), BN, ReLU&CONV(1,4,1), Sigmoid\\
        CONVT(3,4,3), Tanh&Output: 1\\
        Output: (32, 32, 3)&\\
    \end{tabular}
  \end{center}
\end{table}

\subsection{Model architecture for experiments on convolution network depth}
\label{sec:architectureconvdepthexp}
For evaluating the impact of convolution network depth on CNN and FCC-GAN architecture, we increased the number of convolution layers in the generator and discriminator of CNN models from 4 to 7. The CNN model architecture is shown in Table~\ref{tab:cnnmodel7convlayer}. The FCC-GAN models for this experiment were obtained by replacing the first (last) convolution layer in the generator (discriminator) with three (four) FC layers. Table~\ref{tab:fccganmodel7convlayer} shows the FCC-GAN architecture with 7 convolution layers. 

\begin{table}[h!]
\small
  \begin{center}
    \caption{Generator and discriminator network for CNN models with 7 convolution layers}
    \label{tab:cnnmodel7convlayer}
    \begin{tabular}{c|c}
        \multicolumn{2}{c}{} \\ \hline 
        Generator&Discriminator\\ \hline
        Input: Z(100)&Input: (32,32,3)\\
        R(1,1,100)&CONV(64,3,1), BN, LReLU\\
        CONVT(256,4,1), BN, ReLU&CONV(64,4,2), BN, LReLU\\
        CONV(256,3,1), BN, ReLU&CONV(128,3,1), BN, LReLU\\
        CONVT(128,4,2), BN, ReLU&CONV(256,4,2), BN, LReLU\\
        CONV(128,3,1), BN, ReLU&CONV(256,3,1), BN, LReLU\\
        CONVT(64,4,2), BN, ReLU&CONV(256,4,2), BN, LReLU\\
        CONV(64,3,1), BN, ReLU&CONV(1,4,1), Sigmoid\\
        CONVT(3,4,3), Tanh&Output: 1\\
        Output: (32, 32, 3)&\\
    \end{tabular}
  \end{center}
\end{table}

\begin{table}[h!]
\small
  \begin{center}
    \caption{Generator and discriminator network for FCC-GAN models with 7 convolution layers}
    \label{tab:fccganmodel7convlayer}
    \begin{tabular}{c|c}
        \multicolumn{2}{c}{} \\ \hline 
        Generator & Discriminator \\ \hline
        Input: Z(100) & Input: (32,32,3)\\
        FC(64), ReLU & CONV(64,3,1), BN, LReLU \\
        FC(512), ReLU & CONV(64,4,2), BN, LReLU \\
        FC(8192), BN & CONV(128,3,1), BN, LReLU \\
        R(4,4,256) & CONV(128,4,2), BN, LReLU \\
        CONV(256,3,1), BN, ReLU & CONV(256,3,1), BN, LReLU \\
        CONVT(128,4,2), BN, ReLU& CONV(256,4,2), BN, LReLU \\
        CONV(128,3,1), BN, ReLU & Flatten(8192) \\
        CONVT(64,4,2), BN, ReLU & FC(512), LReLU \\
        CONV(64,3,1), BN, ReLU & FC(64), LReLU \\
        CONVT(3,4,3), Tanh & FC(16), LReLU \\
        Output: (32, 32, 3) & FC(1), Sigmoid \\
         & Output: 1 \\
    \end{tabular}
  \end{center}
\end{table}

\subsection{Experimental settings}
In standard GAN training, we used the classic GAN objective function~\cite{gan} for optimization. All models were trained with a batch size of 32. Adam~\cite{adam} optimizer was used with a learning rate of 0.0001 and a decay of 0.00001. For WGAN training, the Wasserstein distance was used as the objective function. All hyper-parameters were kept similar to those described in ~\cite{wgan}: RMSProp optimizer with a learning rate of 0.00005 and a batch size of 64.

\end{document}